\documentclass[letterpaper, 10 pt, conference]{ieeeconf}  

\IEEEoverridecommandlockouts                              

\usepackage{times}
\usepackage{epsfig}
\usepackage{graphicx}
\usepackage{amsmath}
\usepackage{amsfonts} 
\usepackage{hyperref}
\usepackage{cite}

\usepackage{subcaption}
\DeclareCaptionSubType*[Alph]{table}
\DeclareCaptionLabelFormat{mystyle}{Table~\bothIfFirst{#1}{ ̃}#2}
\usepackage{multirow}
\usepackage{array}
\newcolumntype{P}[1]{>{\centering\arraybackslash}p{#1}}

\usepackage{xcolor}
\usepackage{comment}

\title{\LARGE \bf
Don't Forget to Buy Milk: Contextually Aware Grocery Reminder Household Robot
}

\author{Ali Ayub$^{1,a}$, 
Chrystopher L. Nehaniv$^{2,1c}$, and Kerstin Dautenhahn$^{1,2b}$
\thanks{$^{1}$Department of Electrical and Computer Engineering,
        University of Waterloo, Waterloo, ON N2L3G1, Canada
        {$^{a}$\tt\small a9ayub@uwaterloo.ca}, $^{b}$\tt\small kerstin.dautenhahn@uwaterloo.ca}%
\thanks{$^{2}$Department of Systems Design Engineering,
        University of Waterloo, Waterloo, ON N2L3G1, Canada
        {$^{c}$\tt\small chrystopher.nehaniv@uwaterloo.ca}}
}

\begin{document}

\maketitle
\thispagestyle{empty}
\pagestyle{empty}

\begin{abstract}
\label{sec:Abstract}
Assistive robots operating in household environments would require items to be available in the house to perform assistive tasks. However, when these items run out, the assistive robot must remind its user to buy the missing items. In this paper, we present a computational architecture that can allow a robot to learn personalized contextual knowledge of a household through interactions with its user. The architecture can then use the learned knowledge to make predictions about missing items from the household over a long period of time. The architecture integrates state-of-the-art perceptual learning algorithms, cognitive models of memory encoding and learning, a reasoning module for predicting missing items from the household, and a graphical user interface (GUI) to interact with the user. The architecture is integrated with the Fetch mobile manipulator robot and validated in a large indoor environment with multiple contexts and objects. Our experimental results show that the robot can adapt to an environment by learning contextual knowledge through interactions with its user. The robot can also use the learned knowledge to correctly predict missing items over multiple weeks and it is robust against sensory and perceptual errors.
\end{abstract}

\section{Introduction}
\label{sec:introduction}
\noindent
With an increasing ageing population worldwide \cite{iriondo18,fuss20}, extensive research efforts are being dedicated towards developing autonomous robots that can support assistive living for older adults in their homes. Such robots have already started to appear in various roles, such as cleaning robots, caretakers, and home assistants \cite{Matari17,petrecca_how_2018,Saunders16,Koay21,reiser13}. For many of these assistive robots to complete tasks in the household, it would be required that the items needed to complete the tasks are available in the house. For example, consider a robot that can assist with setting up a table for breakfast. If some of the items needed for breakfast (such as milk or cereal) are missing, the robot cannot immediately assist with adequately fulfilling this task. These items could simply be missing from the kitchen, and replacements might be available in a storage location, or they might be completely missing from the household. In such cases, the household robot must direct its assistance towards reminding its user to either buy the missing items or replace them from the storage location. Therefore, in this paper, our goal is to develop a computational architecture that can allow a household assistive robot to keep track of grocery items in the household and remind its user when any of the items are missing.

Most research on grocery related robots has focused on general-purpose commercial robots that either provide assistance in grocery stores \cite{Dworakowski21}, or help deliver groceries to users' homes \cite{Liu21,chivarov21}. Although these robots can assist us with finding groceries in the store and get them delivered to our homes, they cannot help us with the groceries that we might need or items that we might forget to put on our grocery list. For such cases, we need personalized household robots \cite{Dautenhahn04} that understand what grocery items we generally use, and assist us by tracking the groceries in the house and reminding us when some of the items are lacking. 

Research on grocery reminder or recommender systems is limited. Most of the research in this field has been on developing algorithms to recommend new grocery items based on their similarity with the other grocery items bought by the user in the past \cite{Mantha20,Bodike20,Li09}. These systems, however, do not remind users about the groceries that are missing from a household. Further, most of these systems are integrated in smartphone apps. However, these apps are not easily accessible to the older adults (only 20\% of seniors have access to smartphones \cite{cameron20}). Recently, internet of things (IOT) based solutions have also been developed to track grocery items in smart fridges \cite{Miniaoui19,Morris21}. However, these systems can track only a small number of items (only 2 items in \cite{Morris21}), and they also require users to put the correct grocery items in pre-specified bins in the smart fridge. Further, these systems can only track items that are in the fridge, but they cannot predict missing items from other parts of the house. To the best of our knowledge, we know of no other work on developing personalized robots that can assist with detecting regularly used missing groceries from a household and reminding users to buy the missing items. To create assistive robots that can remind about missing groceries in a way that effectively supports diverse users, it is first necessary for the robot to learn the contextual knowledge of the household, i.e.\ items and their contexts/locations within the given house, since different user households can  have different habits and use different items in different contexts. The robot must also be able to reason on the learned knowledge to detect diverse types of missing items/objects from the household contexts and remind its user when habitually used items are missing. 

\begin{figure*}
\centering
\includegraphics[width=0.8\linewidth]{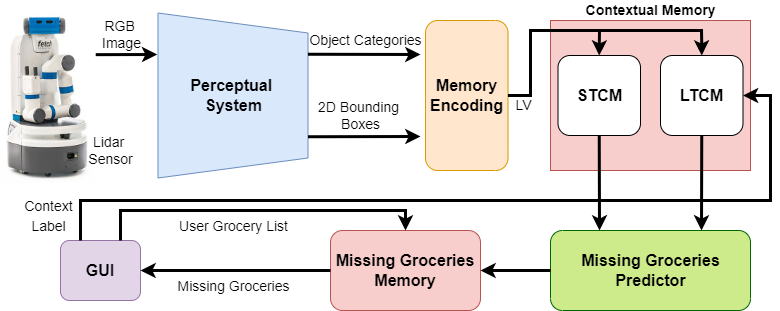}
\caption{\small Complete computational architecture for predicting missing groceries in a household. Sensory input from the Fetch robot is processed through the perceptual system and encoded into a latent variable, which can be used to learn new contextual knowledge in LTCM and store daily experiences in the household in STCM. The prediction module compares the data in STCM and LTCM to predict missing groceries in the household, which can be accessed by the user using a GUI.}
\label{fig:architecture_grocery}
\end{figure*}

In this paper, we develop a cognitively-inspired computational system that can allow a robot to learn contextual knowledge of a household environment from its user, and then use the learned knowledge to make predictions about the missing grocery items in the household over a long period of time. We take inspiration from the dual-memory theory of the mammalian brain, which has considerable experimental support in neuroscience \cite{kitamura2017engrams}. According to this theory, the hippocampus (HC) and the medial prefrontal cortex (mPFC) act as complementary systems, where HC stores dynamic recent experiences for the short-term, and mPFC stores long-term, mostly static memories. Similarly, in our architecture the daily experiences encountered by the robot through its sensors are stored in the short-term contextual memory~(STCM), while the contextual knowledge of the household, grounded in the processed sensory data of the robot, is stored in the long-term contextual memory~(LTCM). The contextual knowledge of the household is learned from interactions with the robot's user, using a graphical user interface (GUI).
The architecture uses a novel technique that compares data in STCM with contextual knowledge in LTCM to predict missing items from the household and stores them in memory. The missing items can be accessed by the user through the GUI. We integrate 
our architecture on the Fetch mobile manipulator robot \cite{Wise16}, and test it in a large indoor space with four different contexts, and 10 different 
household objects. Extensive evaluations in the environment confirm that the robot can correctly predict missing grocery items over a long period of time in the environment. The results also show that the robot is flexible and it can correctly update the missing grocery list if the items get replaced or if they are moved to a different context in the environment. Finally, the robot can also track and suggest if the missing items are available in the storage location.

The remainder of the paper is organized as follows: 
Section \ref{sec:methodology} describes our complete architecture for predicting missing grocery items from a household. Section \ref{sec:experiments} presents empirical evaluations of the system. Section \ref{sec:conclusion} offers conclusions and finally, Section \ref{sec:limitations} discusses limitations of the present work and outlines directions for future research. 

\section{Contextual Memory System for a Grocery Reminder Robot}
\label{sec:methodology}
\noindent Figure \ref{fig:architecture_grocery} shows our complete architecture for a grocery reminder robot. Our architecture can learn the contextual knowledge of the household through user supervision and stores it in LTCM. The architecture can then track the items in the household and predict when any of the items are missing. The missing items can be accessed by the user through a graphical user interface (GUI). The major components of the architecture are described below:

\subsection{Robot's Sensors}
\noindent We use the Fetch mobile manipulator robot for this project \cite{Wise16}.  
Fetch consists of a mobile base and a 7 DOF arm. The robot is equipped with an  RGB camera and a depth sensor for 3D perception of the world around it, and a Lidar sensor to map and detect obstacles in the environment. In this architecture, we use the RGB camera and the Lidar sensor for perception, mapping and navigation in the environment.

\subsection{Perceptual System}
\label{sec:perceptual_system}
\noindent The perceptual system of the architecture takes the RGB image from the robot's sensors, and parses the input data into separate objects. We use the YOLOv2 object detector \cite{Redmon_2016_CVPR} to detect objects in the RGB images. The detected objects are then passed through another convolutional neural network (CNN) classifier to get the object categories. YOLO also provides the object categories for the detected objects. However, the predicted categories are biased towards the classes that YOLO was trained on. For custom household objects it is not possible to correctly predict the object categories from YOLO. For example, in the context kitchen, YOLO classifies many objects incorrectly (Figure \ref{fig:gui_grocery}). Objects banana, and apple have both been incorrectly labeled as orange, cereal has been incorrectly labeled as cup, honey has been incorrectly labeled bottle, and milk is not even detected by YOLO. Therefore, a separate CNN classifier is trained on the custom objects in the household, to correctly predict the object categories. More details about the CNN classifier are in Section \ref{sec:experiments}. 
The perceptual system, thus, parses the input images and outputs the object categories, 
and 2D bounding boxes 
for all the objects in the image.

\subsection{Memory Encoding}
\label{sec:memory_encoding}
\noindent One of the widely accepted tenets of theories of memory encoding in neuroscience is that the firing patterns of a neural network can be considered a point in high-dimensional space \cite{rumelhart1986,anderson1977}. The dimensions of this high-dimensional point (called a \textit{latent variable}) encode the characteristics (features) of the world sensed through our sensors. Many approaches have been proposed in computational neuroscience for encoding the sensory inputs into latent variables (LVs). In this paper, we encode the processed sensory inputs by the perceptual system, using \textit{conceptual spaces} \cite{Douven20}. A Conceptual Space is a metric space in which entities are characterized by quality dimensions. Conceptual spaces have mostly been used for category learning, where the dimensions of latent variables (LVs) in a conceptual space represent the category features. In this paper, we use a conceptual space LV to represent contexts in a household (such as a kitchen), where the features of the LV represent the collection of objects in the context represented by the LV.

\subsection{Short-Term Contextual Memory (STCM)}
\label{sec:stcm}
\noindent
Once an input image is encoded into a latent variable, it is stored in the short-term contextual memory (STCM) of the architecture. The size of STCM is set as a hyper-parameter to allow the architecture to store the encoded images for a certain number of days. After the allocated number of days, the reasoning module (Section \ref{sec:reasoning_module}) processes the data in STCM to find the missing grocery items. Once the data in STCM is processed, it is discarded to make room for more incoming data for the next days. The reason for storing the encoded data for multiple days is that there can be cases when there are items missing from the kitchen because the user might have moved them to a different place. For example, a user could move the cereal box from the kitchen to the dining area, while eating breakfast. In this case, it would be wrong to suggest to buy cereal, just because it is missing from the kitchen. Over multiple days, the robot can either find the item in the dining area or the item could get moved back to the kitchen and the robot can find it there. Thus, by storing the household contextual data for multiple days, the robot can accurately predict what items are missing from the household. 

\subsection{Long-Term Contextual Memory (LTCM)}
\label{sec:ltcm}
\noindent The long-term contextual memory (LTCM) stores the contextual knowledge of the household that the robot operates in. As items in different households can be different, a general purpose semantic architecture, such as concept net \cite{speer17} would not be suitable as it is not grounded in the objects present in the household. Therefore, the robot must learn about the household items and their related contexts through human supervision.

In our architecture, a user can initiate a learning session using a GUI (details in Section \ref{sec:gui}) and provide training examples of household contexts to the robot. The robot captures the contexts as images using its sensors. The perceptual system (Section \ref{sec:perceptual_system}) processes the training images which are then encoded into latent variables (Section \ref{sec:memory_encoding}). To learn the contexts from LVs, we use a neuro-inspired network architecture, termed as SUSTAIN  (Supervised and Unsupervised STratified Adaptive Incremental Network) \cite{love04}. The core algorithm of SUSTAIN is a clustering technique, that starts with a simple solution initializing a single cluster per category. New training data (encoded as LVs) is then compared with the previously learned clusters, where a category prediction failure results in the recruitment of a new cluster. Otherwise, all the clusters compete and the cluster closest to the new training data wins and is updated using the data. Details about SUSTAIN can be found in \cite{love04}. One of the advantages of SUSTAIN is that the clusters learned in the network are not only suitable for category prediction, but also can be compared with new data to make predictions about other feature dimensions.

\subsection{Predicting Missing Groceries}
\label{sec:reasoning_module}
\noindent SUSTAIN learns the categorical information about the household contexts in the form of clusters. It can take an LV (Section \ref{sec:memory_encoding}) as an input and return the predicted context category for the LV by comparing it with the learned clusters. However, in case of predicting missing grocery items (or features from the LV), the objective is different from category prediction. In this case, the model must predict the features that might be missing from an input LV in comparison with the learned contextual clusters of the household. 

To achieve this, we add an extra output layer in the SUSTAIN network, where the output dimensions are the same as the LV dimensions. The output LV represents a collection of objects that are missing from the input LV in comparison with the contextual knowledge stored in LTCM. 
To get this output, we compare the input LV to all the clusters learned by SUSTAIN and find the missing features. Mathematically, let $x$ be an input LV that was stored in STCM and is passed through the SUSTAIN network, consisting of $k$ total clusters for all the contexts in the household. Let $C=\{c_1, c_2, ..., c_k\}$ represent a set of centroids of the $k$ clusters in SUSTAIN. To compare $x$ with all the cluster centroids, we use a slightly modified version of equation (5) in \cite{love04}, as follows:

\begin{equation}
    z_{j}^i = e^{\lambda_j \mu_{ji}}
\end{equation}

\noindent Where, $\mu_{ji}=x(j)-c_{i}(j)$ represents the difference between the $j$th dimension of the input LV $x$ and centroid $c_i$ of the $i$th cluster, $\lambda_{j}$ represents the weight of the SUSTAIN network along dimension $j$, and $z^i$ is an LV that represents 
the activation of cluster $i$ along each dimension of the input LV $x$. In equation (1), if feature dimension $j$ is missing in $x$, $x(j)=0$. Therefore, $\mu_{ji}<0$ and $z_j^i<1$ if $c_i(j)>0$ (feature $j$ is present in cluster $i$). Otherwise, if $x_j>0$, $z_j^i>1$. As all the dimensions in $z^i$ with values greater than 1 show that those dimensions were present in the input LV, we clamp those dimensions and set them to zero. The output LV $z^i$ thus have positive values between 0 and 1, for all the dimensions that are missing in $x$. After finding $z_i$ for all $k$ clusters, we take an average of all these LVs along each dimension to generate the final output LV $v$, where the $j$th dimension of $v$ is determined as follows: $v_j = \frac{1}{k} \sum_{i=1}^k z_j^i$. The output LV $v$ (we term $v$ as prediction LV to differentiate it from the input LV $x$), thus represents the missing features from $x$ in comparison with all the context clusters. 

To find the missing groceries from the household over multiple days, we first pass all the LVs $\{x_{t_1},x_{t_2},...,x_{t_N}\}$ stored in STCM through SUSTAIN to obtain the prediction LVs $\{v_{t_1},v_{t_2},...,v_{t_N}\}$. Then, we multiply all the prediction LVs along each feature dimension to get a final LV which represents missing features from input LVs over multiple days. As a zero in a dimension represents that the item was present in the input LV, multiplying along that dimension over all prediction LVs would generate a zero. Intuitively, this means that an item can get moved to different contexts in the household and the robot can miss it some times, but using the input data over multiple days the architecture will be able to predict that the item was not missing from the household. Further, it is also possible that the robot's sensors or the perception system fail to detect and encode objects from the input images sometimes. However, over multiple days if the objects are correctly detected even once, then the architecture can correctly predict that the objects are not missing from the household. This makes our architecture robust to detection errors. Experimental evaluations on a real robot (Section \ref{sec:experiments}) confirm the robustness of our proposed architecture.

The final output LV is then decoded using the inverse of the procedure in Section \ref{sec:memory_encoding} to get the missing objects from the household. The names/labels of the missing objects are then stored in a separate short-term memory.

\begin{figure}[t]
\centering
\includegraphics[width=1.0\linewidth]{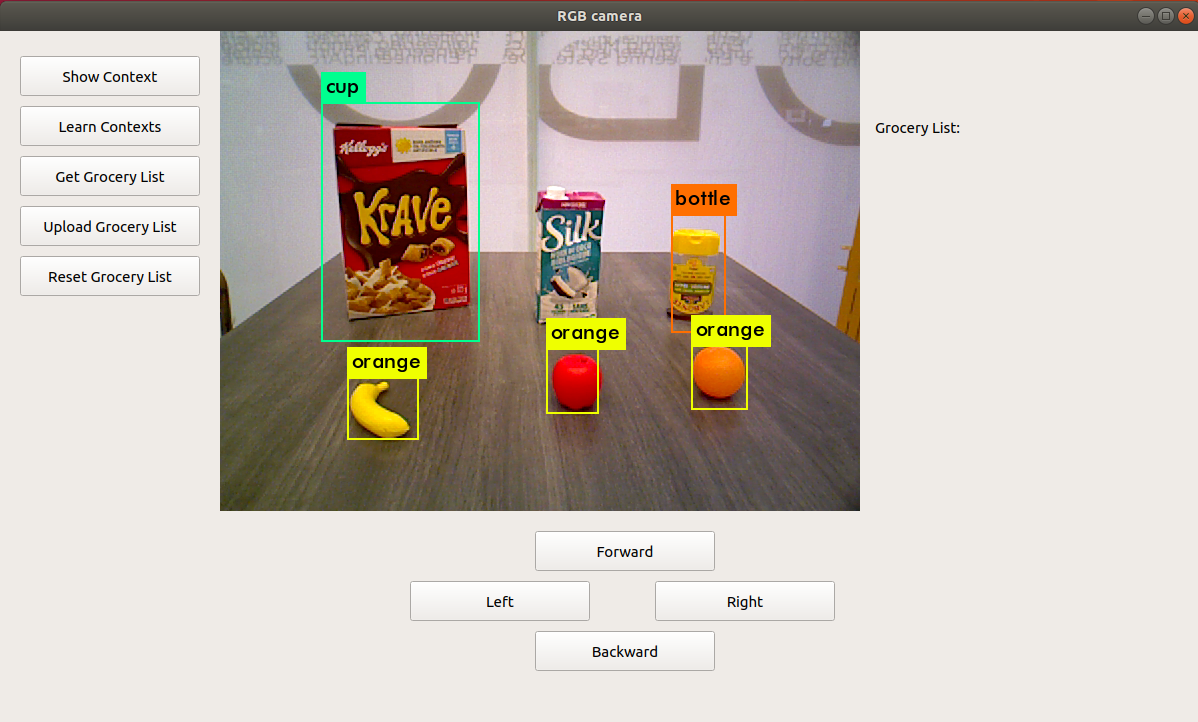}
\caption{\small The graphical user interface (GUI) used to interact with the robot. The head camera output of the Fetch robot wit detected objects through YOLO is in the middle. Missing grocery list is shown on the right of the camera output. Buttons on the left of the camera output can be used to teach contexts to the robot, access the missing grocery list from the robot's memory, upload a new grocery list, and reset the stored grocery list. Buttons at the bottom of the camera output can be used for manual control of the robot.}
\label{fig:gui_grocery}
\end{figure}

\subsection{Graphical User Interface}
\label{sec:gui}
\noindent A simple graphical user interface (GUI) is integrated with the architecture to allow the robot to communicate with the user. The GUI allows the user to initiate a teaching session with the robot where the user can manually move the robot to learn different context locations. The robot captures the data at each context using its sensors and gets the label of the context from the user. The robot then learns and stores the contextual knowledge in LTCM (Section \ref{sec:ltcm}), when the user gives the command (using the button Learn Contexts) to learn from the stored sensory data of the contexts. 

The GUI is also used to share the stored list of missing grocery items with the user. The user can also input a grocery list of their own to the GUI, which is then compared with the missing grocery list in the robot's memory. After comparison, the set of missing grocery items  not in the list provided by the user are displayed to the user. Finally, after replacing the items in the environment, the user can reset the grocery list  Figure~\ref{fig:gui_grocery} shows a picture of the GUI when the robot is in front of the context home\_office.

\section{Experiments}
\label{sec:experiments}
\noindent Here we first describe the experimental setup and the implementation details. We then describe various experiments to evaluate the performance of our proposed architecture for learning different contexts and predicting missing grocery items on a Fetch mobile manipulator robot \cite{Wise16}. For purposes of system evaluation, the experimenters take the role of a user. 

\subsection{Experimental Setup}
\label{sec:experimental_setup}
\begin{figure}[t]
\centering
\includegraphics[width=0.5\linewidth]{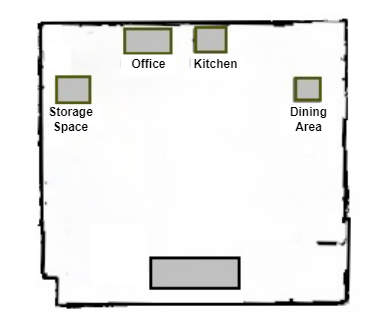}
\caption{\small A slam map of the roamable areas in the indoor environment. Locations of different contexts have been labeled in the map.}
\label{fig:slam_map_env}
\end{figure}

\begin{figure}[t]
\centering
\includegraphics[width=1.0\linewidth]{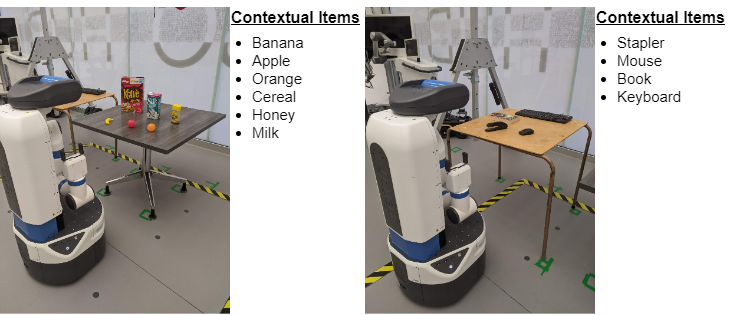}
\caption{\small Fetch robot in contexts kitchen (top) and home\_office (bottom) in the indoor environment.} 
\label{fig:fetch_contexts}
\end{figure}

\noindent We use the Fetch robot for all the experiments. 
We set up various contexts of a household in a large indoor laboratory space (RoboHub at University of Waterloo), with realistic household objects. The indoor space is mapped using the Lidar sensor on the Fetch robot and an existing SLAM algorithm available from Fetch Robotics. Figure \ref{fig:slam_map_env} shows the SLAM map of the environment with four different contexts/locations that are home\_office, dining\_area, kitchen, and storage\_space. Navigation in the environment was achieved using ROS packages provided by Fetch Robotics. Common household items/objects belonging to 10 categories are placed at the appropriate contexts within the environment. Home\_office contains objects belonging to categories book, mouse, keyboard, stapler; kitchen contains objects milk, apple, banana, cereal, orange, honey; storage\_space contains objects cereal, stapler, honey, and the dining\_area does not contain any objects. All context locations contain objects placed on table top environments (Figure \ref{fig:fetch_contexts}), as in most previous works on semantic reasoning systems \cite{chernova2020situated}. Also, note that the total number of objects (10) chosen for our experiments is still much higher than robotics experiments in previous works \cite{chernova2020situated,jiang2019open}.

The RGB camera 
on the Fetch robot 
is used for visual sensing of the environment. RGB images from the camera are passed through the perception module of the architecture which uses YOLOv2 \cite{Redmon_2016_CVPR} to detect and localize objects in the images. YOLO uses pre-trained weights on the COCO image dataset \cite{Lin14}, and it is integrated in our architecture as a ROS package. 
Cropped images of objects detected by YOLO in the input image, are passed through a custom CNN classifier to get the category labels of the objects. For the custom classifier, we first capture a small set of training images for all the 10 object categories used in our experiment. We then pass all the images of the objects through a ResNet-18 \cite{He_2016_CVPR} pre-trained on the ImageNet dataset \cite{Russakovsky15}. The last layer of ResNet-18 is removed to extract $\mathbb{R}^{512 \times 1}$ dimensional features of the images. The object features are then used to train a nearest class mean (NCM) classifier \cite{Mensink13}. For classification of a test object, the object image is first passed through the ResNet-18 feature extractor to get a feature vector for the object. The feature vector is then passed through the NCM classifier to get the category label of the object. For learning contexts in LTCM, we used the cognitively-inspired SUSTAIN network (Section \ref{sec:ltcm}). The implementation details and hyperparameters for the SUSTAIN network were used as in \cite{Ayub_2020_CVPR_Workshops,Ayub2021_ICRAW}. 

For all the experiments, we first collected the data for different context locations in the household using the Fetch robot and the GUI. The stored data is then used by the architecture to learn and store the contextual knowledge of the environment in LTCM. During the test phase, the robot roams around and goes to various locations/contexts within the environment. Note that the robot does not continue to keep roaming the environment for the entire day, as that would be infeasible in a household environment. In a real household, the robot would probably go to different locations in the household when it needs to perform a task in those locations, which can be only a few times a day. Therefore, we manually fix the total number of locations the robot goes to in a single day to 3. The 3 contexts were chosen randomly in each day, where the user provided a command to the robot to visit a randomly chosen context. Further, instead of performing the experiment over multiple days, we ``simulate" the days by letting the robot go to $x \times$3 contexts, where $x$ represents the total number of days. For example, to simulate the roaming operation of the robot over 4 days, we allow the robot to go to 4$\times$3=12 contexts. We set that the user will buy groceries every weekend, therefore, the robot provides the user with a missing grocery list after 6 ``days" of roaming (6$\times$3=18 contexts) in the environment. All the experiments in the paper follow this setup unless stated otherwise. A video about how our system works can be seen at \url{https://youtu.be/oFGil86pBwM}. 

\subsection{Experiment 1: Suggesting Missing Groceries}
\noindent In this experiment, we tested the robot for suggesting missing grocery items every 6 days over 3 weeks of roaming i.e. the robot visited $6 \times 3 \times 3 = 54$ contexts (see Section \ref{sec:experimental_setup} for details). Thus, the robot suggested missing grocery items 3 times over 3 weeks in the indoor environment. We allowed our architecture to store 2 days of contextual data in STCM. In this experiment, we only tested if the robot was able to correctly predict the missing grocery items, store them in its memory and then suggest the missing items at the end of the week. Items that went missing during the week did not reappear in any of the contexts. Further, the storage location was also not used in this experiment, and the robot only checked if the items were present in the other three contexts (kitchen, home\_office, dining\_area) in the environment. Note that the terms week and day were ``simulated" (see details in Section \ref{sec:experimental_setup}).  

Table \ref{tab:missing_groceries} shows the missing grocery list in the robot's memory every 2 days over 3 weeks. In the first week of roaming, item cereal was removed from the kitchen on day 2, item milk was removed on the beginning of day 5 and banana was removed on day 6. On day 7, the robot correctly suggested cereal and milk as missing but failed to predict banana, because banana was removed on day 6 and STM, which has a 2 day buffer, had a record that banana was present in the environment on day 5. The robot also incorrectly predicted apple as missing, even though it was not missing from the environment. The reason was that the perceptual system failed to detect apple when visiting the kitchen context. Further, the robot visited the kitchen only once on days 5 and 6. Therefore, it was not able to correctly predict apple again. Similarly, items milk, stapler, and keyboard were incorrectly predicted as missing after the first 2 days, but they were detected in the environment over the next days of the week. Thus, the robot was able to avoid incorrectly predicting these items as missing. On day 7, all the missing items were replaced in the environment and the robot reset its grocery list. Weeks 2 and 3, followed similar trends with a few differences. In week 2, the robot incorrectly predicted 4 different objects as it mostly visited the dining\_area and did not visit the contexts kitchen and home\_office many times. On the other hand, the robot only predicted one item incorrectly as it was able to visit contexts kitchen and office multiple times over the week. Finally, at the end of week 3, we tested if the robot can take a user uploaded grocery list and suggest missing items. The user grocery list contained following items: banana and mouse. The architecture compared the input list with the missing grocery list in its memory and suggested cereal, milk and apple as the missing items. These results confirm the ability of our architecture to take user grocery lists and predict missing items from the input lists. 

Over the 3 weeks, the robot was able to correctly predict 9 out of 11 items as missing from the indoor environment. The 2 items that it failed to predict as missing, went missing in days 5 and 6 of each week. This confirms that the robot is able to correctly predict all the missing items if they go missing in the earlier days of the week. However, as the robot uses 2 days of data to avoid wrongfully predicting items as missing, it can fail to predict some items that go missing in the later days of the week. Results in the next experiment show more insight into this trade-off. 

Also, the robot incorrectly predicted 6 items as missing, even though they were not missing from the environment. This was caused because of the misdetections and misclassifications by the perceptual system. Over the 3 weeks, the robot encountered 155 objects in 54 visits of the contexts. 68 out of 155 objects (43.8\%) were misclassified or misdetected by the perceptual system. Further, because of the clutter in the environment (shapes on the walls behind the tables, other equipment and robots in the Robohub), the perceptual system also predicted 35 objects (22.6\%) even though they were not a part of the three contexts. Even with state-of-the-art perceptual algorithms, the robot encountered significantly high perceptual errors. However, by visiting the contexts multiple times over the week, the robot was able to avoid incorrectly predicting a large number of objects as missing (only 6 objects (8.8\% were incorrectly predicted as missing out of the 68 incorrectly classified/detected objects). These results confirm the robustness of our architecture against perceptual system errors.

 \begin{table}[t]
\centering
\begin{tabular}{ |P{0.7cm}|P{0.48cm}|P{6.0cm}|}
     \hline
    \textbf{Week} & \textbf{Day} & \textbf{Missing Groceries} \\
     \hline
    \multirow{3}{*}{1}&2 & cereal, \textcolor{red}{milk}, \textcolor{red}{stapler}, \textcolor{red}{keyboard}\\ \cline{2-3}
    &4 & cereal\\ \cline{2-3}
    &6 & cereal, milk, \textcolor{red}{apple}, \textcolor{blue}{banana}\\ \cline{2-3}
    \hline
    \multirow{3}{*}{2}&2 & apple, honey, \textcolor{red}{cereal}\\ \cline{2-3}
    &4 & apple, honey, \textcolor{red}{cereal}, \textcolor{red}{keyboard}, \textcolor{red}{stapler} \textcolor{blue}{orange}\\ 
    \cline{2-3}&6 & apple, honey, orange, \textcolor{red}{cereal}, \textcolor{red}{keyboard}, \textcolor{red}{stapler}, \textcolor{red}{milk}, \textcolor{blue}{banana} \\ \cline{2-3}
    \hline
    \multirow{3}{*}{3}&2 & banana, milk, \textcolor{red}{keyboard}\\ 
    \cline{2-3}&4 & banana, milk, cereal, mouse\\ \cline{2-3}
    &6 & banana, milk, cereal, mouse, \textcolor{red}{apple} \\ \cline{2-3}
    \hline
 \end{tabular}
  \caption{Missing grocery list over the course of 3 ``weeks" in experiment 1. Items that the architecture incorrectly predicted as missing are shown in red, whereas items that the architecture failed to predict as missing are shown in blue.} 
  \label{tab:missing_groceries}
 \end{table}

\subsection{Experiment 2: Reappearing Missing Items}
\noindent In the previous experiment, once an item was removed from the environment, it did not appear again. However, in reality, items can get placed at multiple locations in the house over the course of the week. For example, the user can move cereal from the kitchen to the dining\_area when eating breakfast but they put it back in the kitchen after a couple of days. Further, it is also possible that the user realizes when an item goes missing from the house and buys it during the middle of the week. In such cases, the robot must update its missing grocery list accordingly. Therefore, in this experiment we follow the same setup as in experiment 1, except that we replace some items in the environment after they were removed in the previous days. As in experiment 1, terms week, day, and removal and movement of items were ``simulated" (see details in Section \ref{sec:experimental_setup}).

\begin{table}[t]
\centering
\begin{tabular}{ |P{0.7cm}|P{0.5cm}|P{3.3cm}|P{2.3cm}|}
     \hline
    \textbf{Week} & \textbf{Day} & \textbf{Missing Groceries} & \textbf{Moved/Replaced Items}\\
     \hline
    \multirow{3}{*}{1}&2 & \textcolor{red}{apple} & cereal\\ \cline{2-4}
    &4 & milk, \textcolor{red}{apple}, \textcolor{red}{honey} & cereal \\ \cline{2-4}
    &6 & milk & honey \\ \cline{2-4}
    \hline
    \multirow{3}{*}{2}&2 &  & stapler, apple \\ \cline{2-4}
    &4 & apple, \textcolor{red}{honey}, \textcolor{red}{milk}, \textcolor{red}{cereal} & cereal, banana, book  \\ \cline{2-4}
    &6 & \textcolor{red}{milk}, \textcolor{red}{cereal} & apple, honey, book \\ \cline{2-4}
    \hline
    \multirow{3}{*}{3}&2 & \textcolor{blue}{orange}, \textcolor{blue}{milk} & \\ \cline{2-4}
    &4 & \textcolor{blue}{orange}, \textcolor{blue}{milk}  & cereal, mouse \\ \cline{2-4}
    &6 & orange, \textcolor{red}{cereal, }\textcolor{blue}{stapler} & milk \\ \cline{2-4}
    \hline
 \end{tabular}
  \caption{Missing grocery list over the course of 5 weeks in experiment 2. Items that the architecture incorrectly predicted as missing are shown in red, whereas items that the architecture failed to predict as missing are shown in blue. Items that were moved or later replaced in the environment were updated in the list.} 
  \label{tab:replace_groceries}
 \end{table}

Table \ref{tab:replace_groceries} shows the missing grocery list in the robot's memory every 2 days over 3 weeks. In the first week, item cereal was removed from the kitchen on day 1 and placed in the dining\_area. The robot visited the dining\_area and found cereal there, thus it did not add cereal to the missing grocery list. On day 3, item milk was completely removed from the kitchen. Milk was added in the missing grocery list as it was not found over 2 days in the environment. Finally, on day 6 item honey was moved from the kitchen to the dining\_area. 
On day 7, the robot correctly predicted milk as the only item missing from the environment. Note that after day 2 and day 4, items apple and honey were incorrectly added in the missing grocery list because of perceptual system failures. However, the robot was able to correctly detect these items over the last 2 days (day 5 and 6), and therefore it did not add these items in the missing grocery list.

In week 2, on day 1, item stapler was moved to the kitchen and item apple was moved to the dining\_area. However, the robot did not add these items in the missing grocery list because it was able to find these items in the environment over the first two days. On day 3, apple was removed from the environment, cereal was moved to the dining\_area, and stapler was moved back to the office. On day 4, item cereal was moved back to the kitchen, and both banana and book were moved to the dining\_area. After day 4, the robot correctly added apple in the missing grocery list and did not add banana and book in the missing list. However, it added honey, milk and cereal in the list because of perceptual system failures. On day 5, item banana was removed from the environment, honey was moved to the dining\_area, and apple was replaced in the environment. The robot was able to find apple in the kitchen, and honey in the dining\_area. Therefore, it did not add these items in the missing grocery list. However, it incorrectly predicted milk and cereal as missing because it only visited the kitchen once over the last two days and the perceptual system failed to detect milk and cereal. Similar trend was seen in week 3. 

Over the 3 weeks, the robot was able to predict 3 out of 4 items correctly as missing from the indoor environment. Similar to the previous experiment, the one item (stapler) that the robot failed to predict as missing, went missing in days 5 and 6 of the week. However, over 3 weeks, the robot was able to avoid incorrectly predicting 12 items as missing because they were either moved to different contexts in the environment or they were replaced during the week. These results confirm the ability of our system to avoid false positives when predicting missing items from an environment. 

\subsection{Experiment 3: Storage Space}
\noindent In household environments, it is possible that users buy groceries in bulk and place most of them in storage and only keep the items that are needed in the kitchen. In such cases, the robot must learn about the storage\_space and look in there before adding missing groceries to its list. Therefore, in this experiment, the user teaches the robot about the storage location, so that it does not predict grocery items as missing if the items are present in the storage\_space. To teach the storage\_space, the user takes the robot to the location, use the GUI (Section \ref{sec:gui}) to collect the data for the context, and then ask the robot to learn from the collected data for the context. The indoor environment, thus, had three regular contexts and one special context i.e. the storage\_space. Note that we had to use the same items in the storage\_space that were in the other contexts. However, when the robot goes to the storage location it does not check the other contexts. Therefore, we were able to use the same items in the storage location without affecting our experimental results. 
We performed this experiment over one week only. The rest of the setup was the same as in experiment 1.

Table \ref{tab:storage_location} shows the results for this experiment. On day 1, items milk and cereal were removed from the environment, and no item was removed on day 2. After the first 2 days, the robot visited the first three contexts and predicted both milk and cereal as missing.
The robot then visited the storage\_space to check if the missing items were there, and it was able to confirm that cereal was in storage\_space. Thus, it added only milk in the missing grocery list. The robot also added apple in the missing grocery list because of the perceptual system failure. On day 4, item apple was removed from the environment and honey was moved to the dining\_area. The robot added only apple in the missing grocery list as it was not found in the environment or in the storage\_space. On day 5, item honey was moved back to the kitchen, book was moved to the dining\_area and stapler was removed from the environment. At the end of day 6, the robot did not add book or stapler in the missing grocery list as book was found in the environment and stapler was found in the storage\_space. Overall, the robot was able to predict all the missing items correctly. These results show the flexibility of our architecture to add new special contexts (storage\_space) to make better predictions about missing groceries in the household. The results again confirm the advantage of using multiple days of data to predict about missing groceries. 


\begin{table}[t]
\centering
\begin{tabular}{ |P{0.7cm}|P{0.5cm}|P{3.3cm}|P{2.3cm}|}
     \hline
    \textbf{Week} & \textbf{Day} & \textbf{Missing Groceries} & \textbf{Storage Items}\\
     \hline
    \multirow{3}{*}{1}&2 & milk \textcolor{red}{apple} & cereal\\ \cline{2-4}
    &4 & milk, apple &  \\ \cline{2-4}
    &6 & milk, apple & stapler \\ \cline{2-4}
    \hline
 \end{tabular}
  \caption{Missing grocery list over multiple days in experiment 3. Storage Items show items that were missing from the other three contexts but were found in the storage\_space. Thus, they were not added in the missing grocery list.} 
  \label{tab:storage_location}
 \end{table}

\section{Conclusions}
\label{sec:conclusion}
\noindent Toward realizing the potential for robots to engage in `forms of life'  with humans whereby robots become participants in patterns of daily life of a household \cite{Nehaniv2013},  we presented an architecture for contextual awareness in a household environment, and prediction of missing grocery items from the household by using the learned contextual knowledge. The architecture is integrated on a Fetch mobile manipulator robot using ROS, and tested in a large indoor environment with multiple contexts and objects. Extensive system evaluations demonstrate the ability of our architecture to allow a robot to track and predict missing groceries in the environment over multiple weeks. The results also confirm the robustness of the architecture to avoid wrong predictions about items that are moved within the environment to different contexts.
 We hope such work will lead to designing more effective personalized household robots that can interact with, learn and provide long-term assistance to older adults in their own homes to support independent living. 

\section{Limitations and Future Work}
\label{sec:limitations}
\noindent For purposes of evaluating the robotic system, all the experiments in this research were performed with the experimenters taking the role of users. In the future, we plan to conduct a user study with human participants, where the participants can interact with the robot and use the GUI to teach the robot different contexts in the environment, and access missing grocery items from the robot's memory. Such a user study could provide additional insights into usability, user experience and acceptability of the system.

Our system demonstrated promising results with the chosen hyperparameter settings, such as the size of STCM, number of times a robot visits different contexts in a day, etc. However, it is not clear how changes in hyperparameter values would affect our system. In the future, we hope to conduct further experiments to understand the trade-offs of choosing different hyperparameter values.

Although we consider a large indoor environment with multiple contexts in our experiments, all the contexts had objects placed on table-top environments. In a real household environment, however, all objects will not be placed on the tables or kitchen counters. Instead, objects can also be placed in the cupboard or in the fridge, etc. In such cases, the robot will need to use its arm to open cupboards and the fridge to check for missing items. Designing manipulation algorithms to open cupboards (or a fridge), however, is out of the scope of this work. Another option to solve this problem would be to have sensors placed in the cupboards and the fridge (such as in a smart home) that can track the missing items and send this information to the robot. We hope to explore these limitations in the future versions of our work.





\section*{Acknowledgments}
\noindent This research was undertaken, in part, thanks to funding
from the Canada 150 Research Chairs Program.


{\small
\bibliographystyle{IEEEtran.bst}
\bibliography{main}
}

\end{document}